# Research on Inertial Navigation Technology of Unmanned Aerial Vehicles with Integrated Reinforcement Learning Algorithm

Longcheng Guo[1]　　　　　　　　　　　　　　　　　SalonG2023@outlook.com

[1] Xuhai College of China University of Mining and Technology

## Abstract：

　　With the continuous expansion of unmanned aerial vehicle (UAV) applications[1], traditional inertial navigation technology exhibits significant limitations in complex environments. In this study, we integrate improved reinforcement learning (RL) algorithms to enhance existing unmanned aerial vehicle inertial navigation technology and introduce a modulated mechanism (MM) for adjusting the state of the intelligent agent in an innovative manner[2]. Through interaction with the environment, the intelligent machine can learn more effective navigation strategies[3]. The ultimate goal is to provide a foundation for autonomous navigation of unmanned aerial vehicles during flight and improve navigation accuracy and robustness.

　　We first define appropriate state representation and action space, and then design an adjustment mechanism based on the actions selected by the intelligent agent. The adjustment mechanism outputs the next state and reward value of the agent. Additionally, the adjustment mechanism calculates the error between the adjusted state and the unadjusted state. Furthermore, the intelligent agent stores the acquired experience samples containing states and reward values in a buffer and replays the experiences during each iteration to learn the dynamic characteristics of the environment. We name the improved algorithm as the DQM algorithm. Experimental results demonstrate that the intelligent agent using our proposed algorithm effectively reduces the accumulated errors of inertial navigation in dynamic environments. Although our research provides a basis for achieving autonomous navigation of unmanned aerial vehicles, there is still room for significant optimization. Further research can include testing unmanned aerial vehicles in simulated environments, testing unmanned aerial vehicles in real-world environments, optimizing the design of reward functions, improving the algorithm workflow to enhance convergence speed and performance, and enhancing the algorithm's generalization ability.

It has been proven that by integrating reinforcement learning algorithms, unmanned aerial vehicles can achieve autonomous navigation[4], thereby improving navigation accuracy and robustness in dynamic and changing environments. Therefore, this research plays an important role in promoting the development and application of unmanned aerial vehicle technology.

# 1 Introduction

**1.1 Research Background**

In recent years, the application fields of unmanned aerial vehicles have been expanding, covering various areas, including military operations, logistics and transport, geographic exploration, environmental monitoring, and so on. Navigation has become a focal point for unmanned aerial vehicle usage, as effective and accurate navigation is crucial for safe unmanned aerial vehicle operations and task execution. Common small quadcopter unmanned aerial vehicles mainly rely on inertial navigation to provide position information[5]. The inertial navigation system (INS) is the main component of unmanned aerial vehicle navigation[6]. INS measures the unmanned aerial vehicle's acceleration and angular velocity using an inertial measurement unit (IMU), and calculates the unmanned aerial vehicle's position and attitude through integration. However, traditional INS has significant limitations when facing complex environments and tasks. In dynamic environments, such as strong winds, air interference, or obstacles, the measurements of the IMU may be disturbed, resulting in the accumulation of navigation errors and increased instability.

In order to solve the limitations of traditional INS, researchers in the industry have attempted to correct the results of inertial navigation through filter algorithms, or to combine inertial navigation with computer vision to correct the navigation results. We propose to combine the improved reinforcement learning algorithm with inertial navigation to correct the navigation results. The unmanned aerial vehicle can sense the environment state in real-time and learn the optimal autonomous navigation strategy adapted to the environment through the reinforcement learning algorithm, thereby improving the accuracy and robustness of the navigation.

Our goal is to study the unmanned aerial vehicle's inertial navigation technology combined with reinforcement learning algorithms, and introduce adjustment mechanisms to enable the agents to obtain the maximum reward value in simulation experiments. By integrating the advantages of reinforcement learning, the unmanned aerial vehicle can continuously learn and optimize the navigation strategy during the flight to better adapt to complex navigation scenarios, ultimately achieving autonomous navigation.

**1.2 Study Objective**

Our research aims to combine the improved reinforcement learning algorithm with inertial navigation data to enable the unmanned aerial vehicle to learn more effective navigation strategies and improve navigation accuracy and robustness in dynamic environments. The results of the neural network program running are output in the form of pre-trained models.

Traditional inertial navigation technology has significant performance limitations when facing complex dynamic environments. Therefore, we need to propose a new reinforcement learning algorithm and apply it to navigation tasks to provide a theoretical basis for achieving high-precision autonomous navigation of unmanned aerial vehicles. We hope to provide a new solution for research and practical applications in the field of unmanned aerial vehicle navigation, thus bringing technological breakthroughs to autonomous navigation issues for unmanned aerial vehicles in complex environments.

## 2 Related Work

**2.1 Overview of Unmanned Aerial Vehicle Inertial Navigation Algorithms**

Unmanned aerial vehicle inertial navigation systems (INS) mainly consist of accelerometers, gyroscopes, and magnetometers[7]. Accelerometers are one of the core components of the inertial navigation system, used to measure the acceleration of the unmanned aerial vehicle in three directions. Gyroscopes are used to measure the angular velocity of the unmanned aerial vehicle in three directions[5]. By sensing the unmanned aerial vehicle's rotational motion, gyroscopes can provide continuous measurements of angular velocity. To reduce navigation errors, inertial navigation algorithms generally use filtering algorithms such as Kalman Filter (KF)[8] and Extended Kalman Filter (EKF)[9] to fuse the data from accelerometers and gyroscopes, estimating the unmanned aerial vehicle's position, velocity, attitude, and other information. These filtering algorithms establish state-space and measurement models, combining the system's dynamics and sensor noise characteristics to perform state estimation and prediction.

**2.2 Application of Reinforcement Learning in Navigation Tasks**

Reinforcement learning is a machine learning method that learns the optimal strategy to achieve specific goals through interaction with the environment[10]. In recent years, reinforcement learning has made significant progress in the field of navigation, such as path planning and obstacle avoidance. Reinforcement learning can adapt to dynamic environmental changes in navigation tasks and update strategies through continuous interaction with the environment, thereby improving navigation performance. Reinforcement learning can be used for path planning and control in systems such as unmanned aerial vehicles, autonomous vehicles, and robots. By treating the environmental state as input, reinforcement learning algorithms can learn a policy network to choose the optimal action based on the current state to reach a target location or complete a task. Reinforcement learning can autonomously explore and learn effective path planning strategies based on feedback from reward signals, adapting quickly to changes in complex environments. Furthermore, reinforcement learning can be used to solve obstacle avoidance problems in dynamic environments. In the navigation process, intelligent systems need to perceive and respond to the appearance and movement of obstacles in real-time. Reinforcement learning can learn obstacle avoidance strategies that adapt to dynamic environments through interaction with the environment. The intelligent agent can use reinforcement learning algorithms to select appropriate actions based on the current state, avoiding collisions or finding the optimal path to bypass obstacles.

**2.3 Adjustment Mechanism of our proposed algorithm**

Based on the above content, the reinforcement learning algorithm shows great potential when applied to navigation tasks, but there are also some shortcomings. First, reinforcement learning models usually require a large amount of high-quality data to enable the agent to learn effective navigation strategies. Insufficient dataset quantity or large measurement errors can affect the generalization performance of the model. Second, there are many hyperparameters in reinforcement learning models that need to be adjusted, such as learning rate, reward coefficient, exploration rate, etc. For unmanned aerial vehicle (unmanned aerial vehicle) navigation tasks, the choice of different hyperparameters may have a significant impact on the model's performance. Furthermore, the actual environment of unmanned aerial vehicle navigation tasks is typically complex and dynamic. To achieve the best application effect, the model needs an accurate modeling of the environment.

To address the above shortcomings, we choose the DQN algorithm of reinforcement learning and innovatively introduce an "adjustment mechanism" to adjust the input state of the algorithm, so that it can better interact with the environment. We name the improved algorithm the DQM algorithm, which stands for "Deep Q-network with Modulation" as it combines the adjustment mechanism with the DQN algorithm. In the traditional DQN algorithm, the agent selects its action based on the input state. After selecting the action, the agent performs the action and receives feedback from the environment. The adjustment mechanism focuses on the action execution part. The core of the adjustment mechanism is the "proportional-integral-derivative" control idea. Specifically, in each iteration, the algorithm reads the data from the dataset to obtain the initial state of the agent. Then, the agent needs to choose the appropriate action. The DQM algorithm limits the agent to choose only two actions: adjusting the state or not adjusting the state. If the agent chooses to adjust the state, the initial state is used as the input to the adjustment mechanism in the form of an array. The adjustment mechanism consists of proportional adjustment, integral adjustment, and derivative adjustment, which adjust the input separately and then sum up the results. It should be noted that the state selection of the agent is discrete, while the "proportional-integral-derivative" control idea is generally used to realize the adjustment and control of continuous systems. Therefore, both the integral adjustment and the derivative adjustment should compute the discrete antiderivative/derivative of the input state with respect to time. Furthermore, in combination with unmanned aerial vehicle navigation tasks, the algorithm will calculate the error between the adjusted state and the state calculated by the inertial navigation program, and the error is associated with the reward obtained by the agent. Furthermore, the adjustment mechanism outputs the adjusted state as the next state of the agent, and the agent also receives rewards from the environment. Specifically, the algorithm sets different reward functions so that the agent can maximize the reward value by continuously interacting with the environment and learn the optimal behavioral strategy. During each training process, the agent selects actions to use the learned Q-value function based on the current state; the Q-value function reflects the long-term cumulative reward obtained by taking a certain action in a certain state. If the agent chooses the action corresponding to the highest Q-value, it can maximize the future reward obtained. After obtaining the next state from the adjustment mechanism, the agent updates the Q-value function based on the reward value and the next state. By continuously selecting actions, executing actions, and updating the Q-value function, the agent eventually learns the optimal behavioral strategy. The flowchart and pseudocode of the DQM algorithm is shown in Figure 1 and Figure 2:

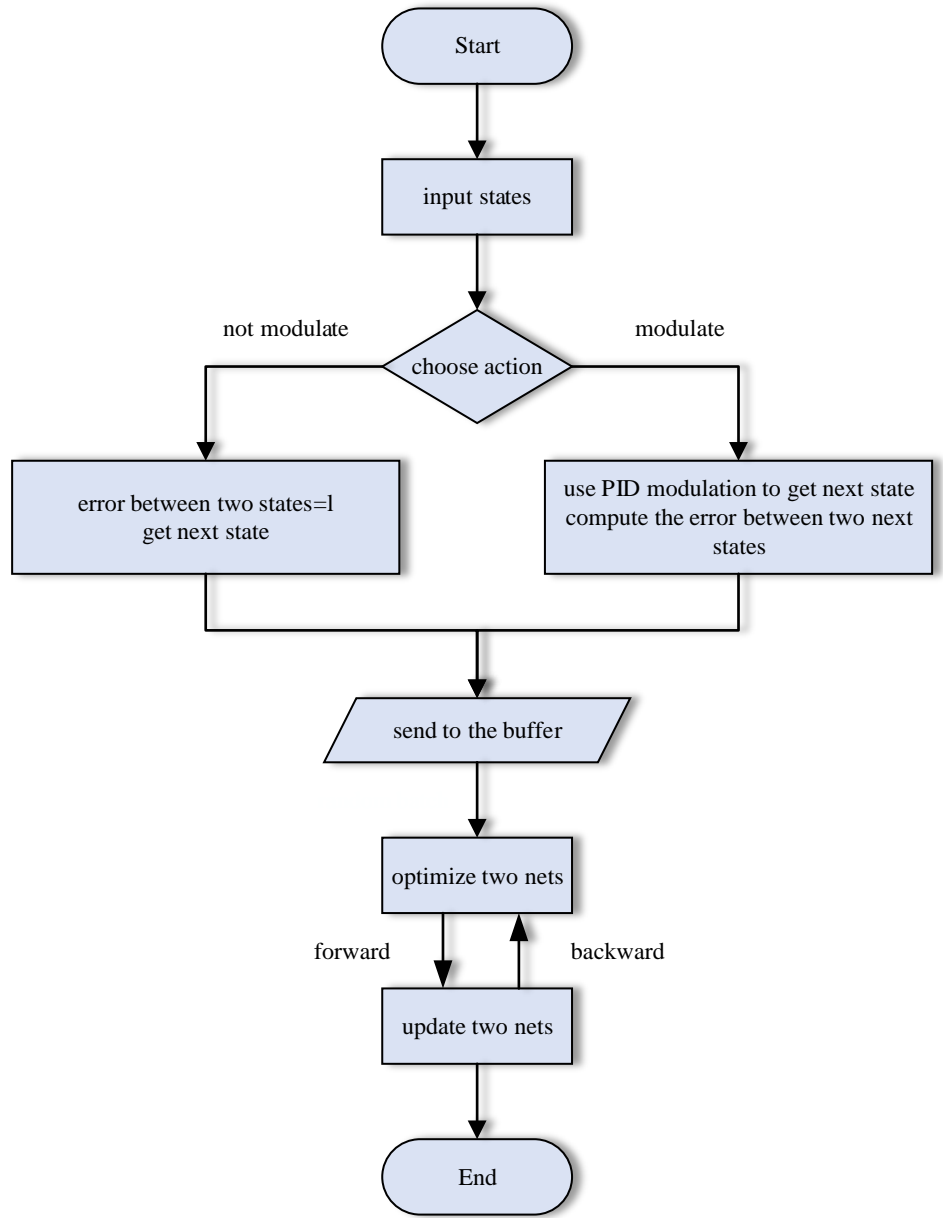

Figure 1: Flowchart of the DQM algorithm

```
Algorithm 1 Deep Q-network with Modulation
Input: Input states
Output: Error between two states and next state
 1: Initialize the input state
 2: if the action is "adjust" then
 3:     Use the PID adjustment mechanism to adjust the input state
 4:     The adjusted state is the next state
 5:     Compute the error between the adjusted state and next state in the
    dataset
 6: else
 7:     Not to adjust the input state
 8:     The next state in the dataset is the next state
 9:     The error between two states equals 1
10: end if
11: Get the error and next state
12: Send it to the buffer of the Neural Network
13: Optimize and update two nets of the Neural Network
```

Figure 2: Pseudocode of the DQM algorithm

The advantages of the DQM algorithm with the introduction of the adjustment mechanism are obvious. Firstly, the improved algorithm simplifies the agent's action selection and reduces the complexity of the agent's action space by setting "adjustment state" and "non-adjustment state" actions, which improves the decision-making efficiency of the agent compared to the original DQN algorithm. Secondly, the improved algorithm associates the error with the reward, which can guide the agent to learn in the direction of reducing the error. This approach will make the agent pay more attention to the effect of state adjustment, thereby allowing the neural network program to converge faster. In addition, the improved algorithm innovatively combines with system control ideas, and the use of proportional-integral-derivative adjustment mechanism allows the agent to adjust the state more flexibly to adapt to complex environments and different task requirements. Finally, based on the proportional-integral-derivative adjustment mechanism, it has a certain degree of interpretability itself; we can adjust the coefficients of proportion, integral, and derivative according to specific requirements and the actual running situation of the program to achieve better state adjustment effects and enhance the adaptability of the algorithm.

**2.4 Application Scenarios of DQM Algorithm**

Applying the DQM algorithm to unmanned aerial vehicle (unmanned aerial vehicle) navigation tasks can fully utilize reinforcement learning's adaptability and dynamic optimization capabilities, providing a foundation for autonomous unmanned aerial vehicle navigation. In recent years, some scholars in the industry have studied and explored the application of reinforcement learning in unmanned aerial vehicle technology. For example, Li Huayuan[11] proposed an adversarial attack reinforcement learning algorithm based on collision risk prediction and opponent modeling for training strategies for unmanned aerial vehicle navigation tasks that balance safety and robustness. Li Yanru[12] and others studied the problem of autonomous planning of unmanned aerial vehicle flight paths based on the DQN algorithm. Luo Jie[13] developed an intelligent flight control system that can simulate and deploy unmanned aerial vehicle reinforcement learning algorithms. Regarding autonomous unmanned aerial vehicle navigation, the industry focuses more on using reinforcement learning algorithms to achieve autonomous planning of unmanned aerial vehicle flight paths and there is less research on combining reinforcement learning algorithms with inertial navigation.

Autonomous unmanned aerial vehicle navigation refers to the unmanned aerial vehicle's ability to independently choose actions and paths based on changes in the environment and task requirements, without human intervention, to complete navigation tasks. Based on the DQM algorithm, unmanned aerial vehicles continuously adjust their behavior and decisions based on the reward signal feedback from the environment to learn the optimal navigation strategy. For example, the algorithm rewards correct execution of tasks or responses to environmental changes while providing less reward for incorrect behavior. Applying the reinforcement learning framework can improve unmanned aerial vehicle's dynamic adaptability. In complex and uncertain environments, unmanned aerial vehicles need to adjust navigation strategies in a timely manner based on changes in the environment and task requirements. Reinforcement learning allows unmanned aerial vehicles to adapt quickly to environmental changes and effectively improve navigation accuracy through real-time perception and learning. In addition, reinforcement learning can help unmanned aerial vehicles achieve environment perception and decision-making capabilities. When unmanned aerial vehicles navigate autonomously in real-world environments, they need to perceive and make decisions based on various environmental information, such as sensor data and map information, in real-time. The computer carried by unmanned aerial vehicles will extract and analyze useful features from perception data and make corresponding decisions based on the extracted features, effectively improving navigation robustness.

# 3 Research Method

## 3.1 State Representation and Action Space Definition

The small unmanned aerial vehicle we are studying has 6 degrees of freedom. We assume $(a_x, a_y, a_z)$ that the acceleration components $\varphi$ of the unmanned aerial vehicle in three directions, $\theta$ roll angle, $\psi$ pitch angle, and yaw angle, and the first-order derivative of the three attitude angles - the angular velocity components of the unmanned aerial vehicle in three directions, are integrated into the dataset that we selected as the state representation for the agent and to provide data support for training neural networks and obtaining pre-trained models. To train the DQM algorithm, we store the state in the form of a numpy array in a CSV file. The state representation is an abstraction and description of the current state of the unmanned aerial vehicle, providing important environmental information for the DQM algorithm. At the same time, the action space contains only two actions: adjusting the state or not adjusting the state. By defining the state representation and action space reasonably, the DQM algorithm can learn effective navigation strategies and make intelligent decisions.

## 3.2 Reward Function Design

The design of the reward function is crucial for the training and learning process of the algorithm. Based on the navigation task goals of the unmanned aerial vehicle, we designed a suitable reward function to guide the unmanned aerial vehicle to learn efficient navigation strategies. The reward function can give different reward signals according to the unmanned aerial vehicle's behavior and state changes, thereby guiding the unmanned aerial vehicle to learn the correct actions and decisions. With a well-designed reward function, reinforcement learning algorithms can quickly converge and learn optimal navigation strategies. Considering that the DQM algorithm associates the rewards obtained by the agent with the error between the two states before and after adjustment, the reward function can be written as a univariate function of the error. When the agent does not adjust the input state, the error between the two states before and after adjustment is 0, and the reward value is set to 1. When the agent chooses to adjust the input state, the reward value is r for a single adjustment, and the error between the two states before and after adjustment is loss. Considering that the reward obtained by the agent increases with the decrease of the error, we introduce a function commonly used in elementary mathematics to represent the inverse correlation between the dependent variable and the independent variable. The function relationship is shown in Table 1.

| Function name | Function expression |
| --- | --- |
| Inverse proportion function | r=1/loss |
| Sigmoid function | r=1/(1+e^(loss)) |
| Inverse of logarithmic function | r=1/ln(loss) |
| Inverse of quadratic function | r=1/(loss^2) |
| Inverse of sine trigonometric function | r=1/sin(loss) |
| Inverse of cosine trigonometric function | r=1/cos(loss) |
| Inverse of tangent trigonometric function | r=1/tan(loss) |

Table 1: Relationship between the reward values obtained by the agent at each sampling point and the state error.

**3.3 Selection of Algorithm Framework at the Bottom Layer**

We chose the DQM algorithm as the framework for unmanned aerial vehicle reinforcement learning. The DQM algorithm is an improvement on the DQN algorithm, which is based on the Q-learning algorithm. By combining deep neural networks, DQM achieves learning and optimization of decision problems in complex environments. During the training process, the DQM algorithm uses experience replay and target networks to solve the problems of sample correlation and unstable target values in traditional Q-learning algorithms. It also uses PID control and reward-error correlation to focus the training of the agent on reducing the errors caused by inertial navigation and provides the model with a certain level of interpretability. As a result, it has achieved significant results in various complex tasks and has wide application prospects.

**3.4 Description of Inertial Navigation Calculation Program**

We have developed an inertial navigation calculation program for a small unmanned aerial vehicle using the Python language. Firstly, the program is designed for an inertial navigation system that points to the geographic north. The origin of the coordinate system is set at an initial longitude of 116.344695283 degrees east, an initial latitude of 39.975172 degrees north, and an initial altitude of 30 meters. We provide initial values for attitude angles and velocities (corresponding to pitch, roll, and yaw angles in degrees, and velocities in the x, y, and z directions in meters per second). The program takes angular velocities and accelerations as inputs and outputs velocities and attitude angles. In particular, the inertial navigation program considers the variation in the acceleration due to gravity on the Earth's surface.

**3.5 Selection of Training Data**

Our training data comes from the EuROC-MAV dataset (referred to as the MAV dataset). The dataset includes visual and inertial data of unmanned aerial vehicles, saved in bag and zip files, which can be used for simulation in the ROS environment as well as training neural networks. It should be noted that the MAV dataset stores data for different task scenarios in separate folders. For the factory scene, the data is stored in the "machine_hall" folder. For indoor scenes, there are two folders "vicon_room1" and "vicon_room2". Additionally, there are folders for calibration data. In the "machine_hall" folder, there are five subfolders, each storing inertial data from different locations. To simplify the task flow, we only use the data from the factory scene. Specifically, we use the data stored in subfolder 1 (MH_01_easy) as the training set for the model and data from subfolder 2 (MH_02_easy) as the validation set.

**3.6 Generation and Optimization of Pretrained Models**

To obtain and optimize the reinforcement learning model, we wrote a neural network program in Python and used the torch library to establish the basic framework of the neural network. After training the neural network with the training set, we plot the graphs and output the pretrained model in the form of a .pth file. Furthermore, the output model is optimized using the data from the validation set[14]. Optimization adjusts the hyperparameters of the model using the data from the validation set to prevent overfitting. Finally, we evaluate the performance of the model on the training set and validation set using evaluation metrics.

# 4 Experimental Design and Result Analysis

4.1 Software and Hardware Environment for the Experiment

To validate the performance of the proposed algorithm in complex environments, we set up the following experimental environment, which provides adjustable parameters and conditions for training the reinforcement learning model, as shown in Table 2.

| Environmental | Content |
|---|---|
| Software Environment | Windows10；Anaconda3；Pytorch1.7.1； |

| | pandas1.0.5；scipy1.10.0； |
|---|---|
| Hardware Environment | CPU: 8 core Xeon 6248R; GPU: RTX3090*1; |
| Storage Space | Memory: 16GB; VRAM: 24GB; |

Table 2: Experimental Environment Description

**4.2 Performance on the Training Set**

The neural network program was developed using the Python language based on the selected data, and it was trained on the training set. Considering that each training iteration on the training set requires the computation of 36,820 sample points (flight data sampled within a 36,820-nanosecond time interval), in order to avoid excessive resource consumption, the number of training iterations was set to 20. The settings for other hyperparameters are referred to in Table 3:

| Name of parameters | Hyperparameter Values |
|---|---|
| Batch Size (BATCH_SIZE) | 32 |
| Learning Rate (LR) | 0.001 |
| Agent Greedy Rate (EPSILON) | 0.9 |
| Reward Coefficient (GAMMA) | 0.9 |
| Target Network Update Frequency (TARGET_REPLACE_ITER) | 100 |
| Memory Capacity (MEMORY_CAPACITY) | 2000 |
| Agent Action Number / Neural Network Output Nodes (N_ACTIONS) | 2 |
| Agent State Number / Neural Network Input Nodes (N_STATES) | 6 |
| Proportional Control Coefficient (kp) | 1.0 |
| Integral Control Coefficient (ki) | 0.5 |
| Differential Control Coefficient (kd) | 0.2 |
| Reward Function | Sigmoid Function |
| State Error Function | Mean Squared Error Function |
| Neural Network Layers | 1 |
| Number of Intermediate Nodes in the Neural | 10 |

| Network | |
|---|---|
| Neural Network Loss Function | Mean Squared Error Function |

Table 3: Hyperparameter Settings for Performance on the Training Set

After training, the total reward value of the agent and the training loss as a function of the number of iterations can be obtained in the following figures:

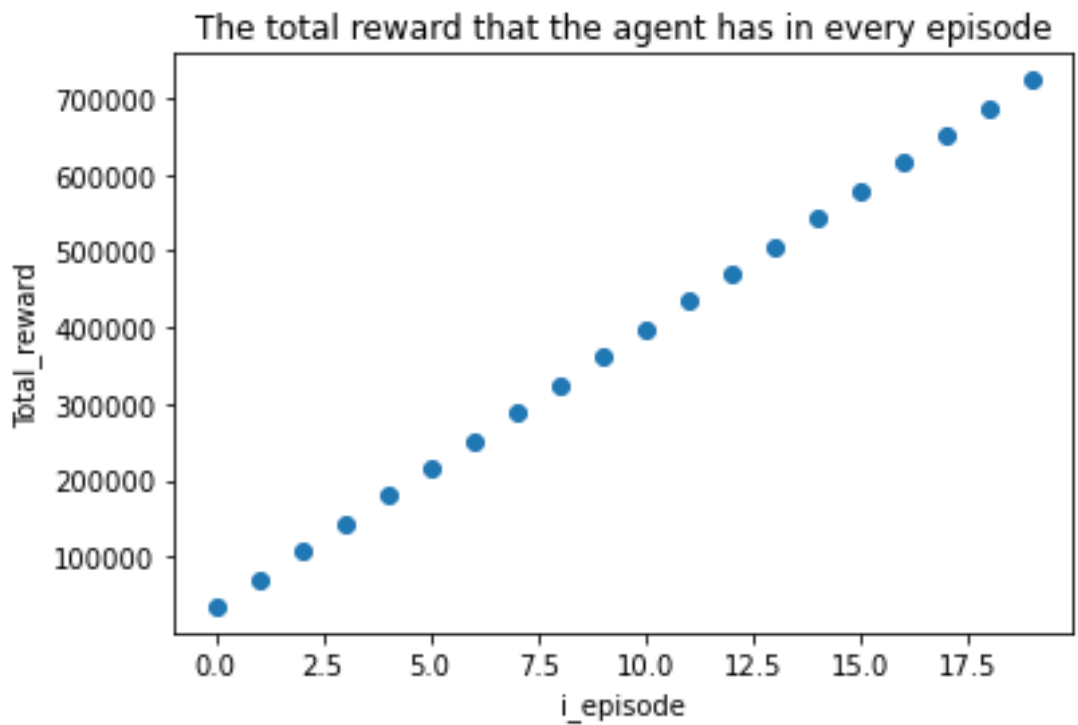

Figure 3: Graph of the Total Reward Value of the Agent as a Function of the Number of Iterations

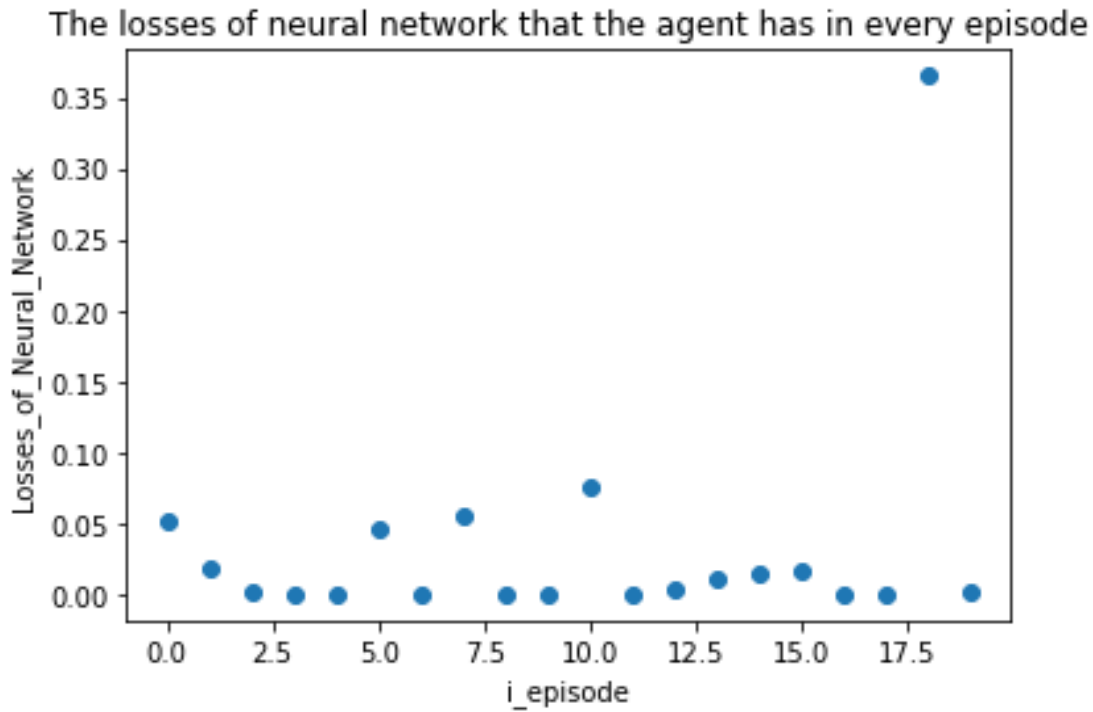

Figure 4: Graph of the Loss Value of the Neural Network as a Function of the Number of Iterations

It is evident from the figures that our model performs well on the training set. As the number of training iterations increases, the total reward value obtained by the agent shows a "linearly increasing" trend. This is consistent with our design idea for the reward function under the DQM algorithm framework, which means that the agent receives a reward regardless of the action it takes, and the reward for adjusting the state is greater. In this case, the training goal of the agent is to maximize the reward value, and the training process will move towards "reducing the error of inertial navigation". In addition, for the training of the neural network, the error between the generated and the target labels needs to be computed at the end of each iteration, which serves as the loss function for the neural network. Our model is no exception, as the error between the evaluation network and the target network is computed at the end; the error eventually converges, proving that the adjustments made to the input state of the agent based on the proportional-integral-derivative (PID) control concept are effective in reducing the error in unmanned aerial vehicle's inertial navigation. However, starting from the third training iteration, the loss value of the neural network fluctuates slightly as the number of iterations increases, which is related to the precision of the data provided by the dataset; once the measurement precision exceeds a certain threshold, the decimal places computed by the program no longer have physical significance, and the loss calculated by the loss function will increase. Therefore, a more reasonable training strategy is to reduce the size of the neural network and the number of training iterations, which not only reduces the computational requirements but also facilitates the deployment of the model on small unmanned aerial vehicles in practical applications.

### 4.3 Performance on the Validation Set

In order to better evaluate the model's generalization ability, the validation set contains data that is similar to but different from the training set, aiming to reflect the performance of the agent in different environments. The validation set consists of 30,400 samples, recording flight data over a period of 30,400 nanoseconds. Under the same values of hyperparameters, after 20 rounds of validation, we can obtain the graph showing the total reward of the agent and the training loss as the number of rounds progresses:

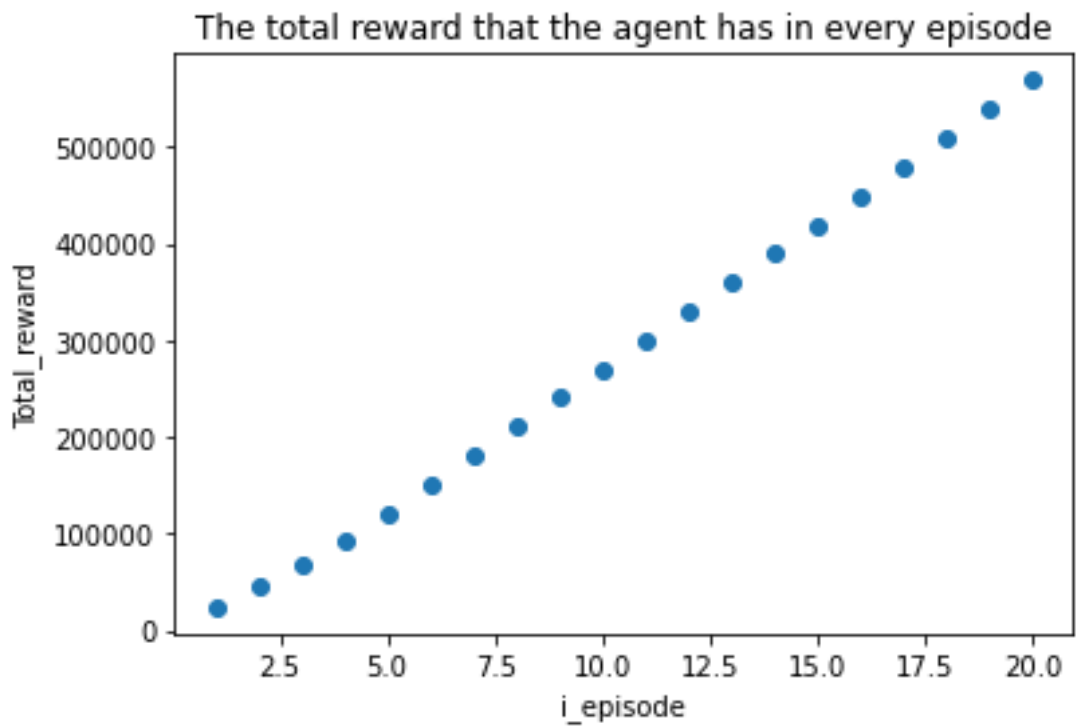

Figure 5: Graph showing the change in total reward of the agent with the number of rounds

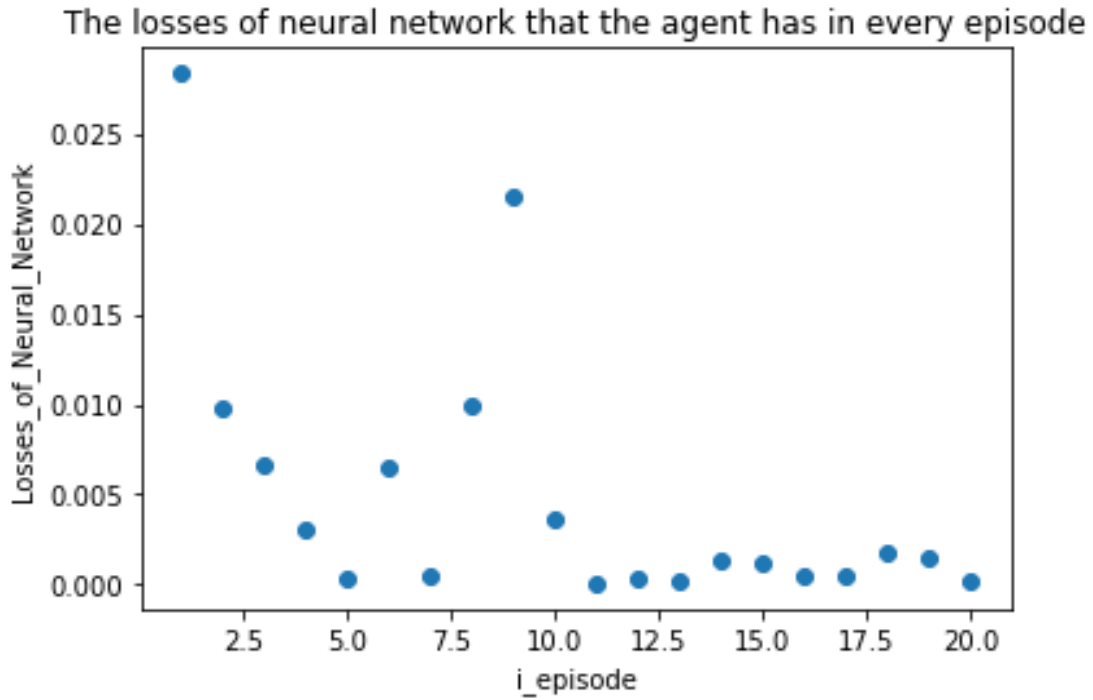

Figure 6: Graph showing the change in loss of the neural network with the number of rounds

Consistent with the performance on the training set, the total reward of the model on the validation set shows a "linearly increasing" trend with the number of rounds, and the model performs better on the validation set. The loss of the neural network for each round shows a trend of "initial decrease and then convergence" as the number of rounds increases, and the loss eventually approaches 0. This fully demonstrates the superiority of the algorithm after introducing the adjustment mechanism, and the model performs exceptionally well on the validation set.

## 5 Discussion and Outlook

### 5.1 Discussion of results

The DQM algorithm has shown significant advantages in this experiment. In comparison, this algorithm can improve the efficiency of agent navigation decisions in complex environments and reduce the errors of inertial navigation. The experimental results confirm the effectiveness of reinforcement learning in the field of unmanned aerial vehicle navigation and provide a new solution to the problem of autonomous unmanned aerial vehicle navigation.

The DQM algorithm we designed performs better in autonomous unmanned aerial vehicle navigation tasks. In this experiment, the intelligent agent trained by us achieved maximized rewards and minimized neural network losses through each round of iteration. The agent focuses more on the goal of reducing "inertial navigation errors" in a single round, and the algorithm has good convergence and high efficiency.

However, our research still has many limitations. Firstly, the adjustment mechanism of the DQM algorithm designed by us requires computing all the sampling points in the dataset, which greatly increases the computational power required to generate the pre-training model. Secondly, the dataset we used has a sampling time unit of nanoseconds, while the actual small unmanned aerial vehicle deployment models often use seconds as the sampling time unit. There are details to be improved in terms of real-time performance and measurement accuracy of the data. Furthermore, our research has not yet revealed the underlying logic behind the adjustment mechanism, and the theoretical rigor on which the algorithm relies may be reduced. Furthermore, the experimental results may be influenced by multiple factors, such as the design of the reward function and the selection of algorithm parameters. Although the reward function we designed performs well in practice, we have not further demonstrated its superiority in this type of problem. Furthermore, the data set size selected in our experiment is small, making it difficult to train a widely applicable pre-training model with strong generalization capability. Our future research can explore the aforementioned issues, such as improving the direction of the adjustment mechanism, optimizing the design of the reward function, evaluating performance in different environmental settings, algorithm adaptability, and deployment issues of the pre-training model, etc.

**5.2 Algorithm Optimization Directions**

Although the experimental results show good performance, there is still a large space for optimization. Here are some possible algorithm optimization directions:

1. Optimization of reward function design: The design of the reward function is crucial for the convergence speed and performance of the algorithm. To further optimize the reward function, one can consider introducing more navigation metrics and objectives, increasing the complexity of the function relationship to more precisely guide the unmanned aerial vehicle to learn ideal navigation strategies.

2. Selection and improvement of the underlying framework: Although the DQN algorithm was used as the underlying framework in this experiment, there are other reinforcement learning algorithms that can be tried, such as Proximal Policy Optimization (PPO), Deep Deterministic Policy Gradient (DDPG), etc. By selecting different algorithms and making improvements, one can try combining different modules and frameworks to ultimately achieve better training results.

3. Fine-tuning of algorithm parameters: The parameter settings of the algorithm have a significant impact on performance. By systematically adjusting algorithm parameters such as learning rate, reward coefficient, etc., the convergence speed and stability of the algorithm can be optimized.

4. Generalization and adaptability of the algorithm: Further research on how to improve the algorithm's generalization and adaptability is an important optimization direction. In practical applications, unmanned aerial vehicles may face different environments and tasks, so the algorithm needs to be able to adapt to different scenarios and requirements. This can be achieved by introducing more training data, increasing the diversity and complexity of the environment, and conducting more rigorous testing and validation.

5. Practical application of the algorithm: How to apply the algorithm to small-scale unmanned aerial vehicle systems in real-world scenarios, taking into account hardware limitations and real-time requirements, is also a research problem that needs to be further explored.

In conclusion, although the unmanned aerial vehicle inertial navigation technology incorporating the DQN algorithm has shown good performance in dataset training, we can still continuously design and optimize the architecture of the algorithm to make it better suited for the real environment of unmanned aerial vehicle flight and provide a foundation for achieving autonomous navigation of unmanned aerial vehicles.

**5.3 Application Prospects**

The fusion of reinforcement learning algorithms and unmanned aerial vehicle inertial navigation technology is a technology with broad application prospects. In the military field, unmanned aerial vehicles are one of the important military equipment, and improving navigation performance and achieving autonomous navigation are crucial for the execution of military tasks. Reinforcement learning-based unmanned aerial vehicle inertial navigation technology can enable unmanned aerial vehicles to better adapt to and perform tasks in complex battlefield environments, such as target reconnaissance and target strikes. In the field of logistics transportation, unmanned aerial vehicles have become an important means of logistics delivery. Reinforcement learning-based unmanned aerial vehicle inertial navigation technology can improve the path planning and obstacle avoidance capabilities of unmanned aerial vehicles in rapidly changing environments, thereby improving the efficiency and safety of logistics transportation. In addition, reinforcement learning-based unmanned aerial vehicle inertial navigation technology can also be applied in the fields of geographic surveying and environmental monitoring. Through the high-precision navigation capability of unmanned aerial vehicles, it can achieve accurate collection of geographical information and real-time acquisition of environmental monitoring data. Therefore, the application prospects of unmanned aerial vehicle inertial navigation technology fused with reinforcement learning algorithms are quite broad.